\documentclass[10pt,twocolumn,letterpaper]{article}

\usepackage[pagenumbers]{cvpr} % To force page numbers, e.g. for an arXiv version

\definecolor{cvprblue}{rgb}{0.21,0.49,0.74}
\usepackage[pagebackref,breaklinks,colorlinks,allcolors=cvprblue]{hyperref}
\usepackage{colortbl}
\usepackage{makecell}
\usepackage{multirow}
\usepackage{pifont}
\usepackage{marvosym}

\def\paperID{6111} % *** Enter the Paper ID here
\def\confName{CVPR}
\def\confYear{2025}

\title{Learning Human Skill Generators at Key-Step Levels}

\author{
Yilu Wu\textsuperscript{1,*} \quad
Chenhui Zhu\textsuperscript{1,*}\quad
Shuai Wang\textsuperscript{1}\quad
Hanlin Wang\textsuperscript{1}\quad\\
Jing Wang\textsuperscript{1}\quad
Zhaoxiang Zhang\textsuperscript{2}\quad
Limin Wang\textsuperscript{1, 3, \Letter }\quad\\
$^1$State Key Laboratory for Novel Software Technology, Nanjing University \\
$^2$State Key Laboratory of Multimodal Artificial Intelligence Systems (MAIS), CASIA \\$^3$ Shanghai AI Lab
}

\begin{document}
\maketitle
\newcommand\blfootnote[1]{%	
  \begingroup
  \renewcommand\thefootnote{}\footnote{#1}%
  \addtocounter{footnote}{-1}%
  \endgroup
}

\begin{abstract}
We are committed to learning human skill generators at key-step levels. The generation of skills is a challenging endeavor, but its successful implementation could greatly facilitate human skill learning and provide more experience for embodied intelligence. Although current video generation models can synthesis simple and atomic human operations, they struggle with human skills due to their complex procedure process. Human skills involve multi-step, long-duration actions and complex scene transitions, so the existing naive auto-regressive methods for synthesizing long videos cannot generate human skills.
To address this, we propose a novel task, the Key-step Skill Generation (KS-Gen), aimed at reducing the complexity of generating human skill videos. Given the initial state and a skill description, the task is to generate video clips of key steps to complete the skill, rather than a full-length video. To support this task, we introduce a carefully curated dataset and define multiple evaluation metrics to assess performance. Considering the complexity of KS-Gen, we propose a new framework for this task. First, a multimodal large language model (MLLM) generates descriptions for key steps using retrieval argument. Subsequently, we use a Key-step Image Generator (KIG) to address the discontinuity between key steps in skill videos. Finally, a video generation model uses these descriptions and key-step images to generate video clips of the key steps with high temporal consistency. We offer a detailed analysis of the results, hoping to provide more insights on human skill generation. All models and data are available at \href{https://github.com/MCG-NJU/KS-Gen}{https://github.com/MCG-NJU/KS-Gen}.
\end{abstract}  
\blfootnote{*~:Equal contribution.}
\blfootnote{\Letter~: Corresponding author (lmwang@nju.edu.cn).}
\section{Introduction}
\label{sec:intro}

\begin{figure}[t]
  \centering
   \includegraphics[width=0.9\linewidth]{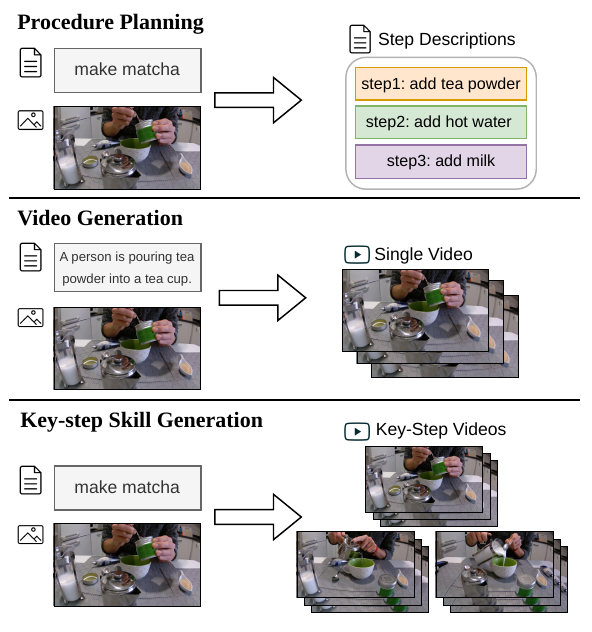}
   \caption{This figure presents three different tasks related to human skill learning. Given an initial image and a text prompt, procedure planning generates a series of steps in textual form. Video generation models can produce a single action video based on detailed text prompts. In contrast, KS-Gen generates multiple key-step videos that complete the skill, using only a simple skill description and image as input.}
   \label{fig:intro}
\end{figure}

Human knowledge can be divided into declarative and procedural knowledge~\cite{Knowledge}. Generative models can now produce diverse and high-quality images~\cite{dalle2,sd} and videos~\cite{SVD,aa,AnimateDiff,latte,dynamicrafter,cogvideox} based on the text description by capturing the declarative knowledge of the real world (e.g., fine-grained visual concepts and temporal dynamics of physical law). 
In addition to capture this descriptive knowledge, we argue that another important and challenging task for generative model is to generate procedural knowledge (i.e., human skill). Human skills~\cite{howto100m,COIN,Survery21} refer to the ability of planning a procedure composed of several key-step actions to accomplish a complex goal. Human skill generator aims to learn the underline distribution of complex human skills, such as how to sow seeds and how to replace batteries. With such a skill generator, humans can learn to perform skills following generated videos, and robots can acquire skill knowledge from synthetic experiences~\cite{gen2act}. 

Given the status of current video generation methods~\cite{SVD,aa,AnimateDiff,latte,dynamicrafter,cogvideox}, human skill generation is extremely challenging because a skill involves multiple steps in the correct order, rather than a simple and atomic action. Additionally, We find that a complete skill video contains many redundant segments, such as repeated actions and numerous scene transitions. We believe these redundant segments have limited generative value and significantly increase the difficulty of generation. Therefore, we propose the Key-step Skill Generation (KS-Gen), a goal-driven and multi-step video generation task of producing a sequence of clips corresponding to the key actions in procedure planning.

As shown in Figure~\ref{fig:intro}, we present some existing tasks related to human skill learning. Procedure planning~\cite{DDN} is among the first to introduce procedural knowledge into video understanding, but regardless of whether the input is an image or text, the output is merely a description of the steps, which is less intuitive than a video. In video generation, some studies~\cite{SVD,aa,dynamicrafter,cogvideox} have attempted to generate human action videos, but most are limited to single-step video generation with highly detailed textual descriptions. Other work~\cite{pandora} has employed auto-regressive methods to generate multi-step, long-duration videos, but these videos typically involve continuous steps without significant changes in object states or scene transitions. Our task, by contrast, aims to generate multiple key-step video segments of a skill process, given an image representing the current state and a skill description. This task is more challenging than previous tasks, as it not only requires step planning but also demands the generation of multiple consistent video segments that capture significant state and scene changes. In real-world applications like instructional cooking videos or product assembly tutorials, videos are rarely filmed in a single continuous take without transitions. Instead, they typically consist of multiple segments to better emphasize key steps. Our task aligns with this practical approach, making it highly relevant to these applications.

One challenge in building the key-step generator is the lack of high-quality skill datasets and suitable evaluation metrics. To generate human activities in real-world environments, we build on existing instructional video datasets, such as COIN~\cite{COIN}, CrossTask~\cite{crosstask} and HT-Step~\cite{htstep}, which cover hundreds of real-world skills, along with Kinetics-400 (K400)~\cite{k400} to enhance action quality. However, the dataset quality varies and presents numerous issues. To address this, we develop a data curation pipeline to improve clip quality and propose diverse evaluation metrics to assess KS-Gen performance. Meanwhile, we propose a new framework for KS-Gen, which consists of the following three main components: (i) \textbf{MLLMs planning}: Based on the provided image and skill description, key-step descriptions are generated utilizing multimodal large language models (MLLMs). To ensure better control over the output, a retrieval-based approach is implemented to optimize the accuracy and sequence of the steps. (ii) \textbf{Key-step image generation}: Since only the initial image is available, subsequent clips lack a first-frame image to serve as a reference. Furthermore, due to the discontinuities and state transitions between the key-step clips, autoregressive generation methods are not well-suited for this task. To address the challenge of generating a consistent sequence of images for each key step, we introduce a novel Key-step Image Generator (KIG). Leveraging the initial image and the step descriptions, KIG generates the first frame for each step, ensuring visual consistency and coherence throughout the key-step images. (iii) \textbf{Video generation}: We generate each key-step clip using our fine-tuned video generation models~\cite{aa,SVD,dynamicrafter,cogvideox}, based on the first frame of each key step and its corresponding text description as prompts.

Our contributions are summarized in three aspects: (i) We take the first step towards human skill generation in the wild at key-step levels (KS-Gen), and build a new benchmark for KS-Gen, including a well-curated dataset and various metrics to assess the performance of skill generators. (ii) We propose a novel framework for KS-Gen that includes three main components. We comprehensively investigate this framework by building several baseline methods and conducting detailed analysis, hoping to provide more insights on human skill generation. (iii) We introduce the Key-step Image Generator (KIG) to address the challenge of generating multiple, non-continuous key-step clips. Experimental results demonstrate that incorporating KIG improves both the quality and consistency of generated images.

\section{Related work}
\label{sec:related work}

\textbf{Learning skill knowledge in instructional videos.} Instructional videos provide intuitive visual examples for learners to acquire skill knowledge. In recent years, a number of datasets for instructional video analysis~\cite{COIN,crosstask,MPII,Youcook,Youcook2,50salads,breakfast,EPIC-KITCHENS,assembly101,howto100m} have been collected in the community. There are various work for instructional video understanding, among which DDN~\cite{DDN} first proposed procedure planning in instructional videos, requiring the model to plan an action sequence from the start state to the end state to simulate different human skill processes. Recently, many work~\cite{plate,p3iv,PDPP,E3P} have attempted this task using transformer or diffusion models. However, the outputs of previous tasks were only action labels or textual descriptions, which cannot intuitively display the entire process. Therefore, we propose learning human skill generators at key-step levels in instructional videos, which intuitively present skills in video form.

\noindent\textbf{Learning real-world generators.} UniSim~\cite{unisim} has shown it is possible to learn a simulator of the real world in response to various action inputs ranging from texts to robot controls. UniPi~\cite{unipi} has showcased the effectiveness of utilizing text-conditioned video generation to represent policies. The setting of VLP~\cite{VLP} is similar to ours but focuses only on robotic scenarios. Although previous tasks cover various scenarios, we consider human operation scenarios to be the most valuable and challenging. Thus, our task focuses on generating human operation videos, encompassing a wider range of actions and skills. Compared to prior tasks, our approach is more challenging due to the abstract nature of skill descriptions, which demand advanced planning method. Additionally, skill videos often involve changes in object states, which makes this task even more difficult.

\noindent\textbf{Video generation model.} With recent advance in diffusion models~\cite{ddpm,ldm}, video generation models~\cite{SVD,aa,AnimateDiff,latte,dynamicrafter} can now synthesize diverse and realistic videos. However, generating long-duration videos remains a significant challenge. Sora~\cite{sora} attempted to generate high-quality, long-duration videos by extending Diffusion Transformers~\cite{dit}. Additionally, Pandora~\cite{pandora}, a hybrid autoregressive-diffusion model, simulates world states by generating videos and supports real-time control with free-text actions. However, since the key-steps in skills are not fully continuous, the autoregressive approach is unsuitable for our task. To address the discontinuity between steps, we introduce an innovative Key-step Image Generation method that autoregressively generates the initial image for each key step, followed by video clip generation. Our method effectively addresses the discontinuity between steps while ensuring consistency across step clips within the same skill.

\section{Key-step skill generation}
Given an initial image and a description of the skill, our key-step skill generation aims to generate a sequence of short clips that depict all key steps involved in performing this skill from the current state. In this section, we introduce our task definition in Sec~\ref{task definition}, and the benchmark for our task, which includes a well-curated dataset in Sec~\ref{bench dataset}, and metrics in Sec~\ref{bench metrics}.

\subsection{Task definition}
\label{task definition}
We define an initial image $I_0$, depicting the starting condition of the skill environment, and a textual description of the skill goal $G$ to achieve, which corresponds to the final status. The output is a series of video clips $\{V_0, V_1,...,V_{n-1}\}$ where each video $V_i$ corresponds to one of the key steps necessary for accomplishing the skill as detailed in $G$. The primary objective is to generate a coherent and logically ordered sequence of video segments that collectively synthesize the skill from initiation to completion, accurately reflecting the transitions and outcomes described in $G$. 

\subsection{Dataset construction}
\label{bench dataset}

\textbf{Data source.} We use COIN~\cite{COIN}, CrossTask~\cite{crosstask} and Ht-step~\cite{htstep} as our raw data, which include annotations of step categories and the corresponding segments. COIN involves 180 skills and 778 steps, including 10,250 videos and 40,000 clips. CrossTask involves 18 skills and 133 steps, including 2,325 videos and 19,079 clips. HT-step involves 433 skills and 4,958 steps, including 16,233 videos and 93,997 segments. To better generate human action videos, we also selected 56,915 videos from 88 categories in Kinetics-400~\cite{k400} for auxiliary training. Since these datasets are curated by different communities, it is necessary to overcome their divergence in granularity and annotation quality, even though they all consist of real-world videos. Previous work~\cite{SVD} has demonstrated that data curation is an essential ingredient for video generation. So we design a comprehensive data curation pipeline.

In the aforementioned datasets, the duration of clips ranges from a minimum of 1 second to a maximum of 300 seconds. Considering that longer videos may include repetitive actions and redundant segments, we divide the original clips into several subclips that can accurately describe the corresponding key steps. Each subclip is kept to not exceed two seconds in length.

We filter subclips based on the following points. (i) \textbf{Scene transition removal:} For each clip, we calculate color histogram features and compute the L1 distance between adjacent frames. Frames where the L1 distance exceeds a set threshold are marked as scene transitions, dividing each clip into subclips without transitions. (ii) \textbf{Ensuring appropriate motion amplitude:} Previous methods~\cite{SVD} removed videos with low optical flow, however, they overlooked camera panning segments. To address this, we propose a template matching method utilizing optical flow magnitude histograms. Using RAFT~\cite{raft}, we compute the histogram of optical flow magnitudes for each subclip, and subclips exhibiting high KL divergence from a predefined template are excluded to ensure appropriate motion amplitude. (iii) \textbf{Clip-description alignment:} To align each subclip with its corresponding step description, we use EVA-CLIP~\cite{evaclip} to extract feature vectors from both video frames and text descriptions. We calculate cosine similarity to select subclips with the highest alignment scores. (iv) \textbf{Reducing text-heavy segments:} We identify and remove segments with high text coverage using CRAFT~\cite{craft}, discarding subclips where the average text area in frames exceeds a set threshold. (v) \textbf{Step description optimization:} Initial step descriptions consist of simple verb-noun phrases, which we enhance using image captioner BLIP-2~\cite{blip2} and video captioner VAST~\cite{VAST}. The descriptions are refined through merging with Llama3-8B~\cite{llama} to produce detailed captions, improving video generation quality.

\begin{figure*}[t]
  \centering
   \includegraphics[width=0.7\linewidth]{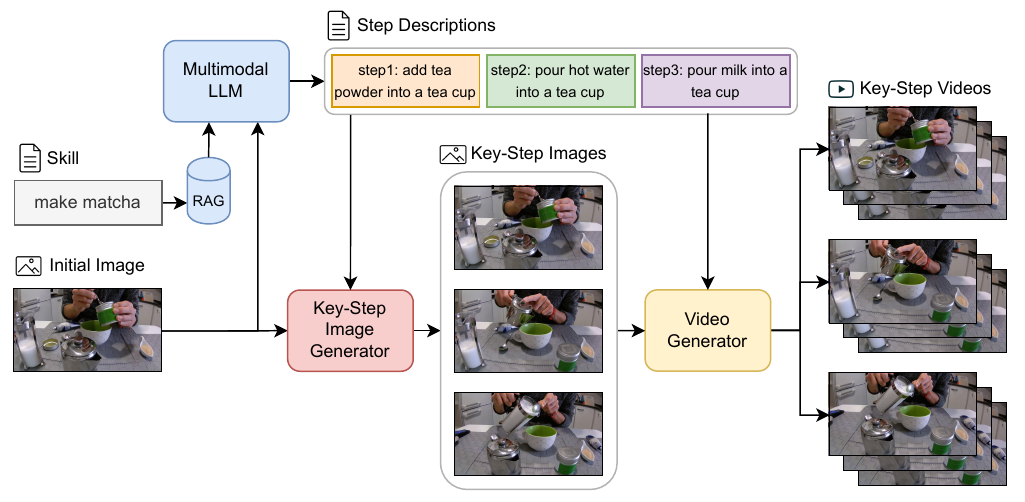}
   \caption{\textbf{Overview of key-step skill generator.} Taking the skill ``make matcha" as an example, this skill includes three key steps. First, based on the given initial image and skill description, we generate detailed descriptions of the three steps through a MLLM using retrieval argument (RAG). (The figure shows the simplified step descriptions.) Then, We input the initial image and step description into the Key-step Image Generation model to generate the first frame of each step. Finally, we use the generated step descriptions as prompts for the video generation model and create video clips corresponding to each of the three key steps, based on the the corresponding key-step images.}
   \label{fig:overview}
\end{figure*}

\textbf{Dataset split.} Through our pipeline, we efficiently curate the above three datasets, culminating in a collection of about about 110,000 subclips for training. Our training set record over 28,500 operational videos from 615 different skills, involving multiple domains such as Housework, Vehicle, Nursing and Care, and more. For testing, we select 557 videos, 1,672 key-steps from the COIN dataset as our test set. Detailed statistics refer to Appendix~\ref{appendix stat}.

\subsection{Metrics}
\label{bench metrics}
To evaluate the performance of our KS-Gen task, we employ a comprehensive set of metrics that collectively assess various aspects of the generated content. These metrics include action similarity, motion dynamics, and overall visual quality. To evaluate the alignment between automatic metrics and human evaluations, we also conducted a user study. For implementation details, please refer to Appendix~\ref{appendix metrics} and Appendix~\ref{user study}.

(i) \textbf{Action score:} We utilize a VideoMAE~\cite{videomae} model that has been fine-tuned on the Something-Something v2~\cite{ssv2} dataset to calculate the cosine similarity of actions between the generated and real videos. This metric helps in understanding how closely the generated actions mirror the intended real activities. (ii) \textbf{CLIP:} We measure the frame-by-frame semantic similarity between generated and real videos by computing the CLIP~\cite{clip} feature similarity. (iii) \textbf{DINO:} Following VBench~\cite{vbench}, we calculate the DINO~\cite{dino} feature similarity to evaluate the semantic consistency of the main objects between the generated and real videos. (iv) \textbf{Motion distance:} The same method previously described for motion score calculation is used here to measure the motion distance between generated and real clips. This ensures that the generated videos exhibit similar motion dynamics as the real actions. (v) \textbf{Fréchet Inception Distance (FID):} FID~\cite{FID} evaluates the quality of generated images by comparing the distribution of generated single frames to real frames from the dataset. Lower FID scores indicate better quality and greater similarity to the original dataset's visual properties. (vi) \textbf{Fréchet Video Distance (FVD):} FVD~\cite{FVD} measures the distance between the distribution of generated videos and real video clips. It is akin to the FID but adapted for videos.

\section{Method}

Considering the complexity of human skill generation in the wild, our framework consists of three main components. As shown in Figure~\ref{fig:overview}, we first generate detailed descriptions of key steps using multimodal large language models (MLLMs), denoted as $\{D_0, D_1,...,D_{n-1}\}$, which is a training-free stage. Additionally, we use retrieval argument to optimize the output of the MLLM. Then we use the initial image $I_0$ and the corresponding descriptions $\{D_0, D_1,...,D_{n-1}\}$ to generate key step clips $\{V_0, V_1,...,V_{n-1}\}$. Since only the initial image is available, subsequent clips lack a first-frame image to serve as a reference. We use a novel Key-step Image Generation~(KIG) model to generate images $\{I_1, I_2,...,I_{n-1}\}$. Finally, we utilize a video generation model to generate key-step clips based on the generated key-step images and descriptions. Below, we provide a detailed explanation of how to utilize MLLMs as planners, along with the process for generating key-step images and videos. 

\subsection{Using MLLMs as a step planner}
\label{sec: planning}
We use the multimodal processing ability of MLLMs~\cite{gpt-4o,gpt-4o-mini,claude} generate step-by-step descriptions $\{D_0, D_1,...,D_{n-1}\}$ for executing the given skills $G$ based on the initial image $I_0$. Having been exposed to massive datasets, MLLMs naturally acquires a variety of skills. Providing only images and skills as inputs to MLLMs for planning may lead to outputs that vary widely. The complexity of planned steps can vary, ranging from highly intricate to overly simplistic. To address this, we propose a retrieval-augmented approach to effectively control the output of MLLMs. Specifically, we build a skill database based on COIN and CrossTask, which includes the names of each skill and corresponding reference step sequences. During the inference process with MLLMs, we first compute the textual similarity between the current skill and all skills in the database, selecting the top three detailed skills and their reference step sequences as examples for MLLMs. We provide detailed prompts in Appendix~\ref{appendix llm}.

We evaluate the quantitative results under a close-set setting, where the skills during inference are all previously seen in the skill database, and we pre-define a set of possible steps and the number of steps need to generate. Acknowledging that a close-set setting may limit the planning capabilities of MLLMs, we also demonstrate qualitative results under an open-set condition. For skills not previously encountered, we do not pre-define a possible set of steps or the number of steps, retaining only the retrieval enhancement, allowing MLLMs to still infer reasonable sequences. We try four different MLLMs~\cite{gpt-4o,gpt-4o-mini,claude} as our planner, and considering the accuracy of generation and the speed of inference, we ultimately choose ChatGPT-4o-latest model as our step planner.

\subsection{Key-step image generation}
\label{image generation}

\begin{figure}[t]
  \centering
   \includegraphics[width=1.0\linewidth]{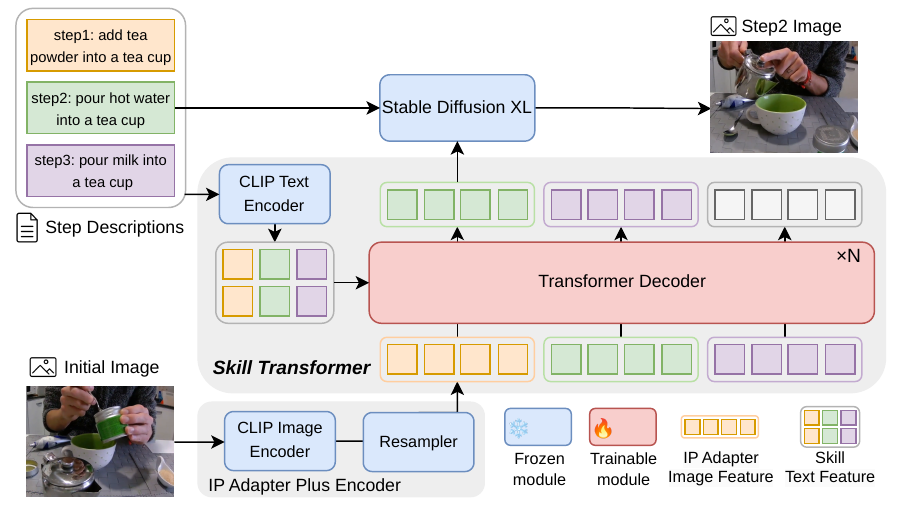}
   \caption{\textbf{Key-step Image Generation.} The input consists of an initial image and step descriptions, from which features are extracted using the IP-Adapter image encoder and CLIP text encoder, respectively. These image and text features are fed into a multi-layer Transformer decoder to autoregressively generate the image features for subsequent clips. The predicted features are then injected into Stable Diffusion XL with IP-Adapter to produce the images.}
   \label{fig:kig}
   \vspace{-10pt}
\end{figure}

To address the challenge of missing $\{I_1, I_2,...,I_{n-1}\}$ and and the inability to use autoregressive methods due to the lack of continuity between key-step clips, we propose a novel Key-step Image Generation (KIG) model. As depicted in Figure~\ref{fig:kig}, KIG consists of two main components: a Skill Transformer and an image generator equipped with IP-Adapter~\cite{ip}. KIG accepts the initial image $I_0$ along with step-wise descriptions $\{D_0, D_1, ..., D_{n-1}\}$ as input. Firstly, a pretrained IP-Adapter~\cite{ip} plus encoder is employed to extract fine-grained image features, denoted as $f_0$, from the initial image $I_0$. Simultaneously, the corresponding textual descriptions of each step are embedded using a CLIP~\cite{clip} Text Encoder, translating these descriptions into skill-specific textual features. Both the image and textual features are integrated and processed through a causal Transformer decoder, which we utilize to predict the IP-Adapter image features for each of the subsequent steps in this skill.
The predicted image features are then passed through the Stable Diffusion XL~\cite{sdxl} model with IP-Adapter to synthesize the corresponding high-quality image for each key-step, ensuring semantic consistency across all key-steps.

During training, we leverage Teacher Forcing by providing the ground-truth features as input to the Transformer. To optimize the model, we employ two MSE losses, one to minimizes the discrepancy between the predicted features and the ground truth feature, and another to ensure the predicted tokens remain consistent with the initial image features $f_0$.
During inference, we autoregressively generate the image features of the remaining key-steps, $\{f_1,f_2,...f_{n-1}\}$. The predicted features are fused with $f_0$ and injected into the existing image model to obtain $\{I_1, I_2,...,I_{n-1}\}$. Notably, our approach does not require fine-tuning of the image generation model, greatly improving training efficiency without sacrificing performance. A complete training process demands only 5 GPU hours and approximately 11GB of VRAM. We use the Stable Diffusion XL (SDXL) as our image generation model and also present results from other image models in the experimental section. 

\subsection{Key-step clips generation}
\label{video generation}
We fine-tune the AnimateAnything (AA)~\cite{aa}, Stable Video Diffusion (SVD)~\cite{SVD}, and DynamiCrafter (DC)~\cite{dynamicrafter} models. Additionally, we evaluate the zero-shot performance of the CogVideoX-5b-I2V~\cite{cogvideox} model. For the AnimateAnything model, which is a text-conditional image-to-video diffusion model, we adhere to the methodology described in the original paper and specifically fine-tune certain layers. In the case of the latent video diffusion model, SVD, since only the image-to-video model parameters are available, we replace the UNet’s conditioning from CLIP image features to CLIP text features and fine-tune only the down blocks and middle block of the UNet. For the DC model, we follow the methodology outlined in the paper and fine-tune all spatial blocks. Furthermore, we report the results for both the DC-512 and DC-1024 model variants. Due to computational constraints, we only evaluate the zero-shot performance of the CogVideoX model.

\section{Experiment}
\label{sec:exp}
Our experiment section first analyze the results of different generators under a standard setting, leveraging the planning capabilities of MLLMs combined with key-step image generator and video generation models to produce key-step clips. We demonstrate the effectiveness of our novel framework across different video models, in comparison to baseline. Subsequently, we conduct detailed ablation studies, including performing multiple ablations on the image and video generation models and planning with various MLLMs. Finally, we display various visualization results of the skill generator under both close-set and open-set settings.

\subsection{Evaluation on key-step skill generators}

\begin{table}[t]
\centering
\resizebox{1\linewidth}{!}{
\begin{tabular}{lcccccc}
\hline
\textbf{Model} & \textbf{Action$\uparrow$} & \textbf{CLIP$\uparrow$} & \textbf{DINO$\uparrow$} & \textbf{Motion$\downarrow$} & \textbf{FVD$\downarrow$} & \textbf{FID$\downarrow$} \\ \hline
DC~\cite{dynamicrafter}+Gen~\cite{genhowto} & 40.8 & 74.9 & 56.3 & 2.17 & 144 & 22.7 \\
DC~\cite{dynamicrafter}+SDXL~\cite{sdxl} & 40.2	& 77.5 & 58.6 & 2.17 & 127 & 21.4 \\
DC~\cite{dynamicrafter}+Ours & 40.8 & 78.1 & 58.9 & \textbf{2.12} & \textbf{119} & \textbf{20.5} \\ \hline
Cog~\cite{cogvideox}+Gen~\cite{genhowto} & 42.9 & 75.3 & 58.7 & 2.87 & 188 & 26.5 \\
Cog~\cite{cogvideox}+SDXL~\cite{sdxl} & 42.5 & 77.6 & 60.4 & 3.06 & 138 & 24.4 \\
Cog~\cite{cogvideox}+Ours & \textbf{43.2} & \textbf{78.4} & \textbf{61.1} & 3.11 & 127 & 22.2 \\ \hline
\end{tabular}}
\caption{Evaluation on baseline methods. We evaluate different video models under different settings based on step descriptions generated by a multimodal LLM. ``Gen" refers to using the GenHowTo model, ``SDXL" refers to using SDXL model with IP-Adapter plus, and ``Ours" denotes using the KIG model to generate key-step images.}
\label{tab:overview}
\vspace{-15pt}
\end{table}

% We constructed a baseline method using GenHowTo~\cite{genhowto} as the image generation model. Specifically, GenHowTo is a generative model designed to produce object state changes in instructional video, conditioned on both images and text. In this work, we directly use GenHowTo to generate key-step images $\{I_1,I_2,...I{n-1}\}$ without fine-tuning. However, Genhowto is not entirely zero-shot, as it is trained on instructional videos~\cite{crosstask,changeit}. We evaluated the generation results using DynamiCrafter (DC)~\cite{dynamicrafter} and CogVideoX-5b-I2V (Cog)~\cite{cogvideox} as video models. While Dynamicrafter was fine-tuned on the training data, CogVideoX was not fine-tuned due to resource and time constraints. As shown in Table~\ref{tab:overview}, our method outperforms GenHowTo across nearly all metrics, with significant improvements observed in FVD and FID. Specifically, when using CogVideoX as the video model, FVD is reduced by approximately 50, and there are notable gains in both CLIP and DINO metrics. These results indicate that our approach leads to substantial improvements in video generation, enhancing both foreground and background quality. Furthermore, these results are obtained under a standard setting, where only the initial image and skill description are provided during the inference process. Ablation experiments on both the video and image models will be presented in the following sections.

We constructed a baseline method using GenHowTo~\cite{genhowto} and SDXL~\cite{sdxl} model with IP-Adapter~\cite{ip} plus as the image generation model. Specifically, GenHowTo is a generative model designed to produce object state changes in instructional video, conditioned on both images and text. In this work, we directly use GenHowTo to generate key-step images $\{I_1,I_2,...I_{n-1}\}$ without fine-tuning. However, GenHowTo is not entirely zero-shot, as it is trained on instructional videos~\cite{crosstask,changeit}. We evaluated the generation results using DynamiCrafter (DC)~\cite{dynamicrafter} and CogVideoX (Cog)~\cite{cogvideox} as video models. While DynamiCrafter was fine-tuned on the training data, CogVideoX was not fine-tuned due to resource and time constraints. As shown in Table~\ref{tab:overview}, our method outperforms GenHowTo and SDXL across nearly all metrics, with significant improvements observed in FVD and FID. Specifically, when using CogVideoX as the video model, FVD is reduced by approximately 50 than GenHowTo, and there are notable improvements in both CLIP and DINO metrics. These results indicate that our approach enhances both foreground and background quality. Furthermore, these results are obtained under a standard setting, where only the initial image and skill description are provided during the inference process.

\subsection{Ablations}

\begin{table}[t]
\centering
\resizebox{0.6\linewidth}{!}{
\begin{tabular}{lccc}
\hline
\textbf{Model} & \textbf{CLIP$\uparrow$}               & \textbf{DINO$\uparrow$}               & \textbf{FID$\downarrow$}                \\ \hline
GenHowTo~\cite{genhowto}       & 66.8                        & 46.4                        & 67.7 \\ \hline
Kolors~\cite{kolors}         & 68.1                        & 46.5                        & 67.6 \\
Kolors ft.~\cite{kolors}     & 69.0   & 46.5  &  67.2                            \\
Kolors~\cite{kolors}+ST.  & 69.9                        & 47.5                        & 67.1 \\ \hline
SDXL~\cite{sdxl}           & 72.5 & 49.5 & 64.8 \\
SDXL ft.\cite{sdxl}       &72.8 & 49.4 & 64.3 \\
\rowcolor[rgb]{0.9,0.9,0.9}SDXL~\cite{sdxl}+ST.    & \textbf{74.1}                        & \textbf{50.4}                        & \textbf{61.1}                        \\ \hline
\end{tabular}}
\caption{Ablations on image generation models. Since Kolors and SDXL were originally designed for text-to-image generation, we use the IP-Adapter to inject the initial image as an image condition into both models. ``ft.” denotes the image model is fine-tuned, while ``+ST.” indicates that image features are passed through our trained Skill Transformer before being injected into the image model. The default choice is colored in \colorbox[rgb]{0.9,0.9,0.9}{gray}.}
\label{tab:ablation image}
\end{table}

\textbf{Image generation model.} To validate the effectiveness of our Key-step Image Generator (KIG), we evaluated its impact on Kolors~\cite{kolors} and SDXL~\cite{sdxl}, using CLIP, DINO, and FID metrics to assess image generation quality. We also choose the GenHowTo model, which is pretrained on instructional videos, as a baseline method. The Kolors and SDXL we used are modified with the IP-Adapter~\cite{ip} plus, as these models were initially designed purely for text-to-image generation. As shown in Table~\ref{tab:ablation image}, SDXL outperforms Kolors and GenHowTo in baseline generation quality. Integrating our Skill Transformer with both Kolors and SDXL yield notable improvements. Additionally, we find that combining the Skill Transformer with non-fine-tuned image models yields superior results compared to using fine-tuned image models alone. This demonstrates that our approach is both highly effective and efficient. To show the robustness of our model, we employ step descriptions generated by an our MLLM, which may contain inaccuracies. Our novel approach effectively enhances model robustness, even with imperfect input descriptions.

\begin{table}[t]
\centering
\resizebox{1\linewidth}{!}{
\begin{tabular}{lccccccc}
\hline
\textbf{Model} & \textbf{Act.$\uparrow$} & \textbf{CLIP$\uparrow$} & \textbf{DINO$\uparrow$} & \textbf{Mot.$\downarrow$} & \textbf{FVD$\downarrow$}  & \textbf{FID$\downarrow$} & \textbf{Time} \\ \hline
DC$_{1024}$ zs.~\cite{dynamicrafter}      & 49.6            & 90.0          & 82.9          & 4.26            & 108.5         & 8.14         & 1.56          \\
Cog zs.~\cite{cogvideox}         & \textbf{54.9}   & 90.1          & \textbf{84.6} & 2.36            & \textbf{63.7} & 8.73         & 3.29          \\ \hline
AA ft. ~\cite{aa}           & 45.7            & 85.4          & 80.3          & 2.67            & 152.6         & 15.8         & 0.14          \\
SVD ft. ~\cite{SVD}          & 45.3            & 86.1          & 78.1          & 2.07            & 162.6         & 15.2         & \textbf{0.12} \\
DC$_{512}$ ft.~\cite{dynamicrafter}         &50.5                &87.7               &79.0               &1.75                 &  75.2             & 8.85             & 0.41          \\
DC$_{1024}$ ft. ~\cite{dynamicrafter}        & 52.9            & \textbf{90.6} & 83.5          & \textbf{1.55}   & 66.3          & \textbf{7.82} & 1.55          \\ \hline
\end{tabular}}
\caption{Ablations on video generation models. During inference, the model is provided with ground truth initial frames and LLM-fused step descriptions. ``zs." stands for zero-shot models, ``ft." indicates fine-tuned models. DC$_{512}$ refers to the generated videos have a resolution of $512\times$320. DC$_{1024}$ refers to the generated videos have a resolution of $1024\times$576.}
\label{tab:ablation video}
\vspace{-10pt}
\end{table}

\textbf{Video generation model.} We evaluated the performance of four different video generation models. To isolate the effect of the video models, we provided each model with the ground-truth first frame of each clip and the step description during generation. As shown in Figure~\ref{tab:ablation video}, we fine-tuned models AA~\cite{aa}, SVD~\cite{SVD}, and DC~\cite{dynamicrafter}, with DC tested at two different resolutions with two model variants. Due to resource limitations and the slower training and inference speeds of CogVideoX~\cite{cogvideox}, we evaluated it in a zero-shot setting only. Results indicate that DC achieves comparable performance to CogVideoX with half the generation time, showing particularly strong improvements in motion metrics, and DC has a significantly smaller model size than CogVideoX. Additionally, we tested DC in a zero-shot setting and observed that fine-tuning the video models led to a notable enhancement in generation quality.

\begin{table}[t]
\centering
\resizebox{1\linewidth}{!}{
\begin{tabular}{lccccccc}
\hline
\textbf{Filter} & \textbf{Action$\uparrow$} & \textbf{CLIP$\uparrow$} & \textbf{DINO$\uparrow$} & \textbf{Motion$\downarrow$} & \textbf{FVD$\downarrow$}  & \textbf{FID$\downarrow$}   \\ \hline
None            & 50.4            & 86.2          & 75.8          & 2.84            & 98.3          & 9.60               \\
+Scene         & 50.8            & 87.6          & 78.3          & 2.34            & 90.0          & 9.21             \\
+Motion        & \textbf{52.2}   & 87.8          & 78.8          & \textbf{1.92}   & \textbf{82.0} & 8.93        \\
+Semtantic     & 51.9            & 88.6          & 80.4          & 1.98            & 86.1          & \textbf{8.86}  \\
\rowcolor[rgb]{0.9,0.9,0.9}+Text          & 52.0            & \textbf{88.6} & \textbf{80.5} & 1.95            & 87.1          & 8.94           \\ \hline
\end{tabular}}
\caption{Ablations on data curation pipeline. We evaluate the performance of the Dynamictafter model under different data curation pipeline. During inference, the model is provided with ground truth initial frames and LLM-fused step descriptions. The generated videos have a resolution of 256$\times$384, consisted of 16 frames, and are generated at 8 fps. In the table, ``+" indicates the addition of the corresponding setting to the previous row. The default choice is colored in \colorbox[rgb]{0.9,0.9,0.9}{gray}.}
\label{tab:ablation data curation}
\vspace{-10pt}
\end{table}

\textbf{Data curation.} The quality of training data is crucial for video generation models, and one of our main contributions is proposing a data curation pipeline that enhances the quality of video clips. Our pipeline mainly consists of four aspects, and we compare the most basic training results, which do not include data processing, with results progressively incorporating different data curation methods. As shown in the Table~\ref{tab:ablation data curation}, removing scene transitions in clips makes the generated videos smoother, leading to improvements in all metrics. The newly proposed method of histogram template matching for optical flow magnitude effectively enhances the motion score, reducing it from 2.34 to 1.92, which convincingly demonstrates the efficacy of this approach. Additionally, handling motion also improves other metrics. Considering that clips and text aligned with semantics are more beneficial for training generative models, we further incorporated semantic processing, which achieves the best results on the FID metric. Training with video segments that contain excessive text can lead to the sudden appearance of text in generated videos, so we removed samples that contained a lot of text. As shown in the last line, by incorporating text processing, we achieved the best results in both CLIP and DINO metrics. Overall, our data curation efforts lead to significant improvements across all metrics.

\begin{table}[t]
\centering
\resizebox{1\linewidth}{!}{
\begin{tabular}{lcccccc}
\hline
\textbf{Prompt} & \textbf{Action$\uparrow$} & \textbf{CLIP$\uparrow$} & \textbf{DINO$\uparrow$} & \textbf{Motion$\downarrow$} & \textbf{FVD$\downarrow$}  & \textbf{FID$\downarrow$}  \\ \hline
Verb-noun       & 51.9            & 88.3          & 79.9          & 1.95            & 85.3          & 9.04          \\
Video           & 51.5            & 88.4          & 80.2          & \textbf{1.88}   & 87.3          & 9.00          \\
Image           & 51.1            & 87.9          & 79.3          & 1.94            & \textbf{83.9} & 9.07          \\
\rowcolor[rgb]{0.9,0.9,0.9} Fusion          & \textbf{52.0}   & \textbf{88.6} & \textbf{80.5} & 1.95            & 87.1          & \textbf{8.94} \\ \hline
\end{tabular}}
\caption{Ablations on step description. We also evaluate the performance of DC model with different step description. The generated videos have a resolution of 256$\times$384. ``Fusion” indicates using LLM to merge multiple descriptions, yielding the best training results. The default choice is colored in \colorbox[rgb]{0.9,0.9,0.9}{gray}. }
\label{tab:ablation prompt}
\end{table}

\textbf{Step Description.} We evaluate video generation models trained with different text prompts. During inference, we uniformly use the ground truth image as the first frame for each clip and the ground truth LLM-fusion description as the prompt. As shown in the Table~\ref{tab:ablation prompt}, model training with LLM-fusion descriptions achieves the best results in four metrics. However, model trained with video captions slightly lead in motion score, and model train with image captions achieve the best FVD. Considering all metircs, we select the fusion description.

\begin{table}[]
\centering
\resizebox{0.6\linewidth}{!}{
\begin{tabular}{lcc}
\hline
\textbf{Method}            & \textbf{SR$\uparrow$} & \textbf{Acc$\uparrow$} \\ \hline
GPT-4o          & 11.3        & 28.5         \\
GPT-4o-mini     & 1.80         & 15.4        \\
Claude 3.5 Sonnet& 15.6       & 33.2        \\
ChatGPT-4o-latest  & 17.1       & 37.7        \\
\rowcolor[rgb]{0.9,0.9,0.9} ChatGPT-4o-latest +RAG                       & \textbf{50.5}       & \textbf{65.0}        \\ \hline
\end{tabular}}
\caption{Ablations on MLLMs planning. We experiment with different MLLMs, where “+RAG” indicates the addition of retrieval-augmented generation. The default choice is colored in \colorbox[rgb]{0.9,0.9,0.9}{gray}.}
\label{tab:ablation llm}
\vspace{-15pt}
\end{table}

\textbf{Multimodal LLM planning.} To verify the accuracy of multimodal LLMs planning sequence, we follow the metrics of procedure planning. As shown in Section~\ref{sec: planning}, we evaluate the step sequences under a close-set setting, along with Success rate (SR) and Accuracy (Acc). SR is the strictest criterion, requiring the sequence to match the ground truth exactly, whereas Acc only requires individual steps to be correct. As shown in Table~\ref{tab:ablation llm}, we test the metrics for GPT-4o~\cite{gpt-4o}, GPT-4o-mini~\cite{gpt-4o-mini}, Claude 3.5 Sonnet~\cite{claude} and ChatGPT-4o-latest~\cite{gpt-4o} without retrieval argument. The lightweight GPT-4o-mini model provides fast inference speed but performs less accurately than other models. ChatGPT-4o-latest yields the best planning results, achieving the highest success and accuracy rates, with a success rate of 17\%. However, a success rate below 20\% highlights the limitations of current multimodal LLMs in directly handling procedure planning tasks, even when selectable steps are provided. With the addition of retrieval argument, there is a significant improvement in both SR and Acc, indicating that the skill database effectively guides the LLM in sequencing planning. We apply retrieval-augmented generation to the ChatGPT-4o-latest model, which serves as our final result.

\subsection{Visualizations}

\begin{figure}[t]
  \centering
   \includegraphics[width=1.0\linewidth]{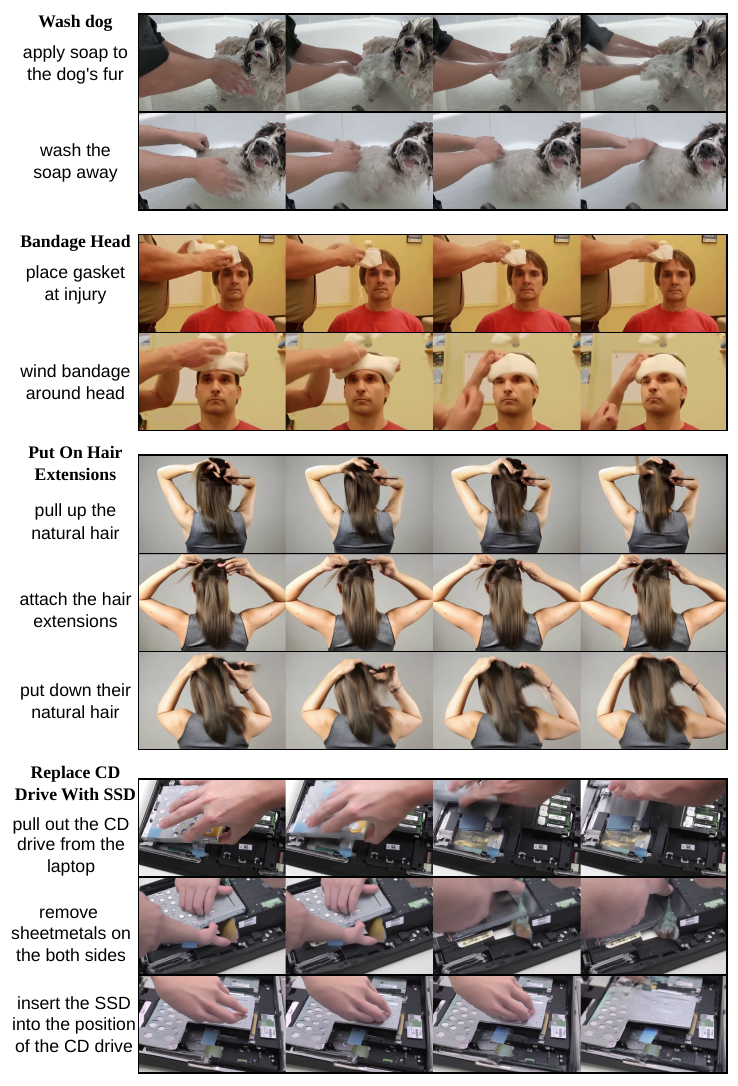}
   \caption{The visualization of the skill generator.}
   \label{fig:vis}
   \vspace{-20pt}
\end{figure}

\begin{figure}[t]
  \centering
   \includegraphics[width=1.0\linewidth]{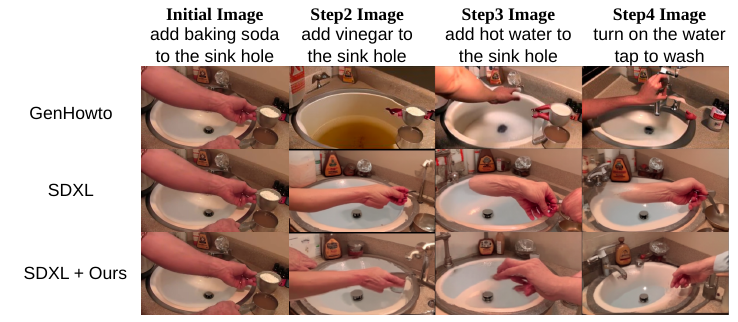}
   \caption{The visulization with different image generation models.}
   \label{fig:ablaimage}
   \vspace{-5pt}
\end{figure}

\begin{figure}[t]
  \centering
   \includegraphics[width=0.97\linewidth]{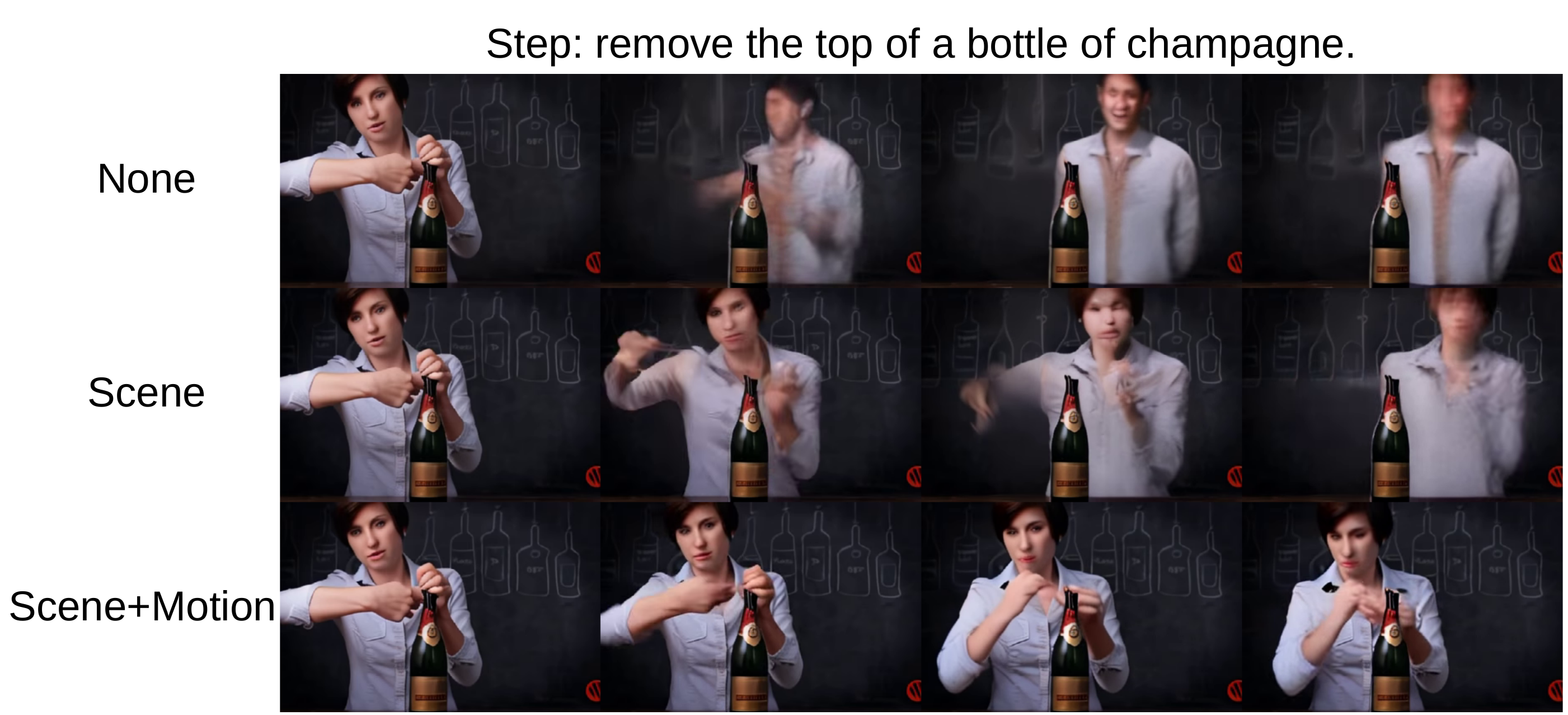}
   \caption{The visualization with different data curation.}
   \label{fig:abladata}
   \vspace{-5pt}
\end{figure}

\begin{figure}[t]
  \centering
   \includegraphics[width=0.97\linewidth]{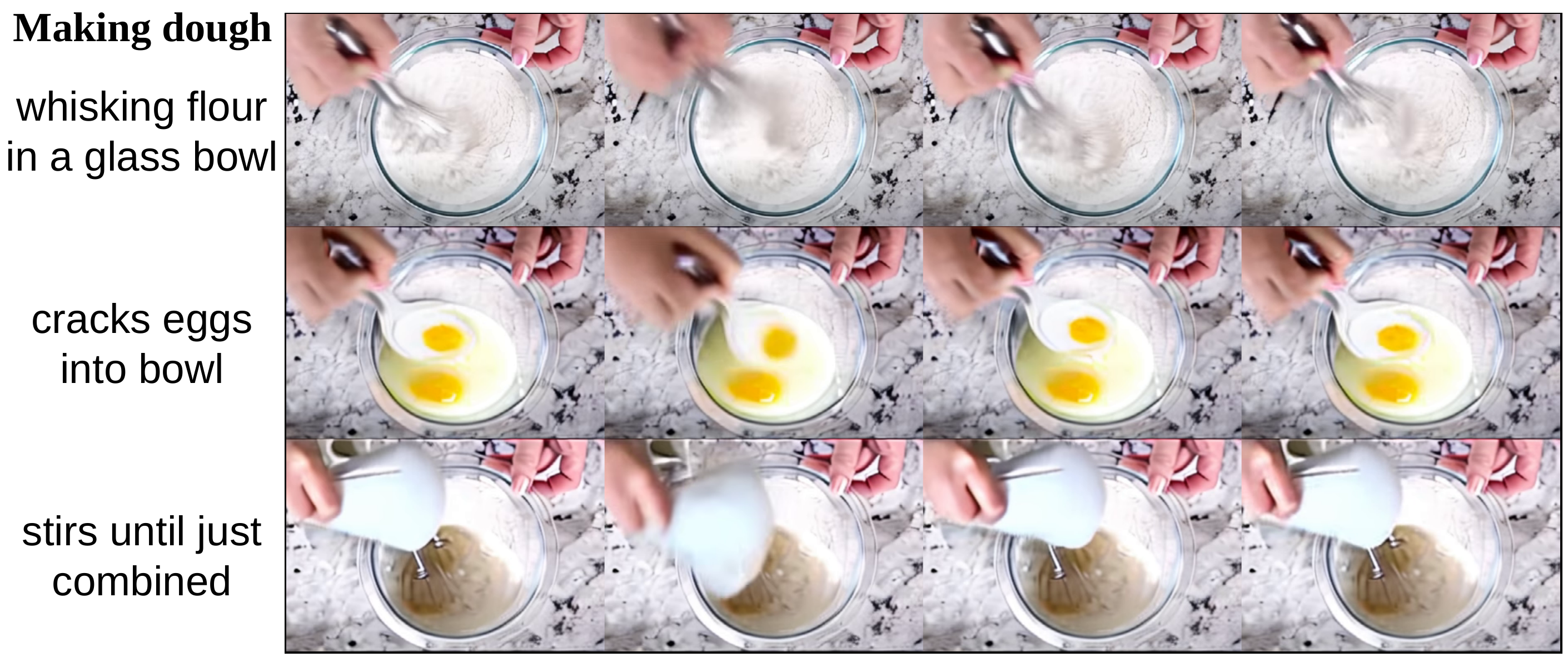}
   \caption{The visualization of unseen skill during training.}
   \label{fig:zeroshot}
   \vspace{-10pt}
\end{figure}

We showcase the visualization results of skill generation. Figure~\ref{fig:vis} displays the outcomes of four different skills generated by CogVideoX~\cite{cogvideox} with our KIG assistance. Additionally, in Figure~\ref{fig:ablaimage}, we present visualization results using different image generation models. The images generated by GenHowTo~\cite{genhowto} exhibit poor consistency and contain errors, such as the soda cup, which should have disappeared in Step 2, remaining in the images. The results using the SDXL~\cite{sdxl} model alone are also suboptimal, with the hand appearing to float in the air in Steps 3 and 4, which is inconsistent with realistic actions. In contrast, our model demonstrates superior consistency and image quality compared to both GenHowTo and SDXL. We present the qualitative analysis results of our data curation, as shown in the Figure~\ref{fig:abladata}. For the first line generated with the raw dataset, the faces of the characters in the video have undergone significant changes, and the action is wrong. By incorporating scene detection, inconsistencies within the clip can be effectively avoided. After we implemented motion control, it is clear that both the consistency of the video and the extent of motion have been greatly improved. Finally, we display the generation results of a skill that have not appeared in the training set as shown in Figure~\ref{fig:zeroshot}. The figures show the simplified step descriptions. More visualizations and detailed descriptions are provided in Appendix~\ref{appendix vis}.

\section{Conclusion and Limitation}

We have introduced a novel task learning human skill generation at the key-step level. For this task, we construct a benchmark and propose a data curation pipeline along with a diverse set of evaluation metrics. We also propose a new framework for key-step skill generation and evaluate various baseline methods. Additionally, the key-step image generator we proposed effectively addresses the challenge of generating non-continuous key-step videos.  However, our framework remains a multi-stage process, limited by current video generation capabilities, and we aim to propose an end-to-end skill generator. We hope that skill generation will not only help humans acquire more skills but also provide more generation data for embodied intelligence, thereby bridging the gap from reality to simulation. Generative models are no longer limited to producing simple actions. The ultimate goal is to generate processes of higher dimensions, as exemplified by our skill generation. We believe KS-Gen will open a promising and meaningful direction for world generation.
%\clearpage
%\setcounter{page}{1}
% \maketitlesupplementary

\newtheorem{sketch_definition}{Definition}[section]
\section*{Appendix}
\appendix

Our appendix includes the following sections: Section~\ref{appendix stat} presents detailed statistics of our dataset. Section~\ref{appendix data} provides an in-depth introduction to our data curation pipeline. Section~\ref{appendix metrics} thoroughly describes the evaluation metrics we employ. Section~\ref{user study} provides the user study we conducted. Section~\ref{appendix llm} explains how we utilize a MLLM to generate step sequences. Section~\ref{appendix kig} details the implementation of the Skill Transformer and the hyperparameters used. Section~\ref{appendix kcg} provides the hyperparameters of the video models we used. Finally, Section~\ref{appendix vis} displays additional visualization results.

\section{Dataset statistics}
\label{appendix stat}
As shown in Table~\ref{tab:statistics}, we display the number of skills, videos and clips include in our dataset. Since K400~\cite{k400} is not an instructional video dataset, it does not contain skill statistics. HT-Step~\cite{htstep} is used exclusively for training the video generation model.

\begin{table}[h]
\centering
\resizebox{1\linewidth}{!}{
\begin{tabular}{lccccc}
\hline
Split & Skills  & Videos & Clips  & Clip Duration & Subclips\\ \hline
Train (COIN+CrossTask) & 182    & 12326       &  58510  & 214 h  & 23997 \\
Train (K400) & -    & -      &  56915 & 154 h  & 22499 \\
Train (HT-Step) & 433  & 16233 &  93997 & 399 h  & 63906 \\
Test  & 133    & 557       &  1672     & 0.89 h & 1672 \\
Total & 615    & 29116   &   133330   &768 h & 112074 \\ \hline
\end{tabular}}
\caption{The statistics of our dataset including COIN, CrossTask, HT-Step and K400.}
\label{tab:statistics}
\end{table}

\section{Dataset curation pipeline}

\label{appendix data}
\textbf{Scene Transition Detection.} For each clip, we compute the 256-dimensional color histogram features for each frame. To detect gradual transitions, we calculate the L1 distance between the histogram features of frames at three intervals: zero frames (adjacent frames), two frames, and four frames. A frame is detected as a scene transition if the L1 distance at any of these intervals exceeds the respective threshold.

\begin{figure}[h]
  \centering
   \includegraphics[width=0.9\linewidth]{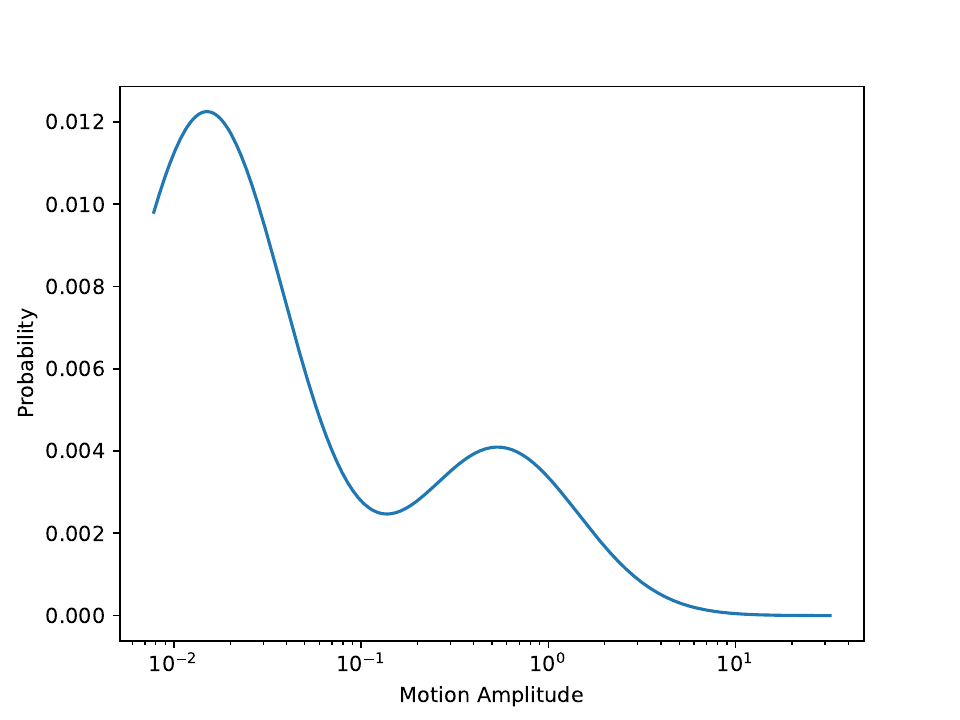}
   \caption{The template of optical flow histogram}
   \label{fig:motion}
\end{figure}

\textbf{Optical Flow Amplitude.} We utilize the RAFT-Large~\cite{raft} model to compute the optical flow between each frame and its adjacent frame for every clip. The optical flow magnitude histograms are then generated, with the magnitude range on a log scale from \(2^{-7}\) to \(2^5\), divided into 256 bins. For each optical flow histogram, we compute the KL divergence from a predefined histogram template. This template, shown in Figure~\ref{fig:motion}, represents a mixture of two Gaussian distributions: the lower mean Gaussian represents static parts of the video (e.g., background), while the higher mean Gaussian represents the motion parts. The ideal video should predominantly contain static parts with some moderate motion parts. After calculating the KL divergence between the template and each frame's histogram, we use a sliding window and a set threshold to identify the most appropriate subclips within each clip while filtering out less suitable subclips.

\textbf{Aligned clip and description.} We employ the EVA01-CLIP-g~\cite{evaclip} model to compute the clip score between each frame and the action descriptions provided in the dataset. Using a sliding window and a set threshold, we identify the most appropriate subclips within each clip, filtering out less suitable subclips.

\textbf{Reducing Text-Heavy Segments.} The CRAFT~\cite{craft} model is used to calculate the proportion of each frame covered by text. Frames with a high proportion of text are excluded based on a predefined threshold.

\textbf{Step Descriptions Enhancement.}
To enhance the quality of step descriptions, we employ the image captioning model BLIP-2-opt-6.4b~\cite{blip2} to generate an image caption every 2 seconds and the video captioning model VAST~\cite{VAST} to produce 10 video captions for each clip. We randomly select three image captions and three video captions, then combine these with the provided step descriptions using the Llama3-8B-instruct model. This fusion process generates finely detailed and accurate captions, further improving the quality of video generation.

\section{Metrics}
\label{appendix metrics}
To measure the quality and similarity of generated videos in comparison to real videos, we employ a set of semantic similarity and video quality metrics. All metrics are computed on 16-frame clips at 8 frames per second (fps) from the test set. For reproducibility, the associated code will be made available. Below are the detailed implementations of each metric:

\textbf{Action Score:} We utilize the VideoMAE-base~\cite{videomae} model fine-tuned on the Something-Something v2~\cite{ssv2} dataset to calculate the cosine similarity between the logits of generated and real video clips. Each video is resized such that its shorter side is 224 pixels, followed by a center crop to 224$\times$224 pixels. Cosine similarity is computed between the 174-dimensional logits vectors of the generated and real videos. For comparison convenience, we scale the similarity by a factor of 100, resulting in an action score ranging from 0 to 100, where higher values indicate better performance.

\textbf{CLIP:} To measure frame-by-frame semantic similarity, we employ the OpenCLIP-ViT-bigG~\cite{openclip} model. Videos are resized to have a shorter side of 224 pixels and then center-cropped to 224$\times$224 pixels. The similarity score between frames of the generated and real videos is scaled by 100 for easier comparison, yielding an image score that ranges from 0 to 100, with higher scores indicating better visual content correspondence.

\textbf{DINO:} Following VBench~\cite{vbench}, we calculate the DINO feature similarity with DINOv2-large~\cite{dinov2} model to evaluate the semantic consistency of the main objects between the generated and real videos. Similar to the CLIP metric, the videos are also cropped to 224 and the similarity score is scaled by a factor of 100.

\textbf{Motion Distance:} Motion distance is assessed using the RAFT-Large~\cite{raft} model to calculate optical flow. We then compute the histogram of the optical flow magnitude on a log scale ranging from \(2^{-7}\) to \(2^5\) with 256 bins. The histograms are normalized, and the Kullback-Leibler (KL) divergence between the histograms of the generated and real videos is computed. The motion distance score ranges from 0 to infinity, with lower values indicating closer similarity. The shorter side of the video is resized to 256 pixels.

\textbf{Fréchet Inception Distance (FID):} We employ the FID implementation from `torchmetrics` to compute the FID score between the frames of generated and real videos. Lower FID scores indicate higher quality and greater similarity to the visual properties of the original dataset.

\textbf{Fréchet Video Distance (FVD):} The FVD score is calculated using the torchscript provided by~\cite{StyleGAN}, comparing the distributions of generated and real video clips. Similar to FID, lower FVD scores signify better performance in terms of video quality and resemblance to real videos.

\section{User study}
\label{user study}
To evaluate the alignment between automatic metrics and human preferences, we conducted a comprehensive user study involving 38 participants. The study was divided into two sections, each focusing on a specific aspect of video generation: action accuracy (whether the video accurately and realistically depicts the described actions) and object consistency (whether the state changes of objects in the video are logical and free from irrelevant artifacts). Each section contained 10 examples, with participants required to rank 4 videos per example based on the given criteria. This structured approach generated approximately 4,500 pairwise comparisons, which were used to compute the average win ratio for each model across the two aspects. 

\begin{figure}[t]
  \centering
   \includegraphics[width=1.0\linewidth]{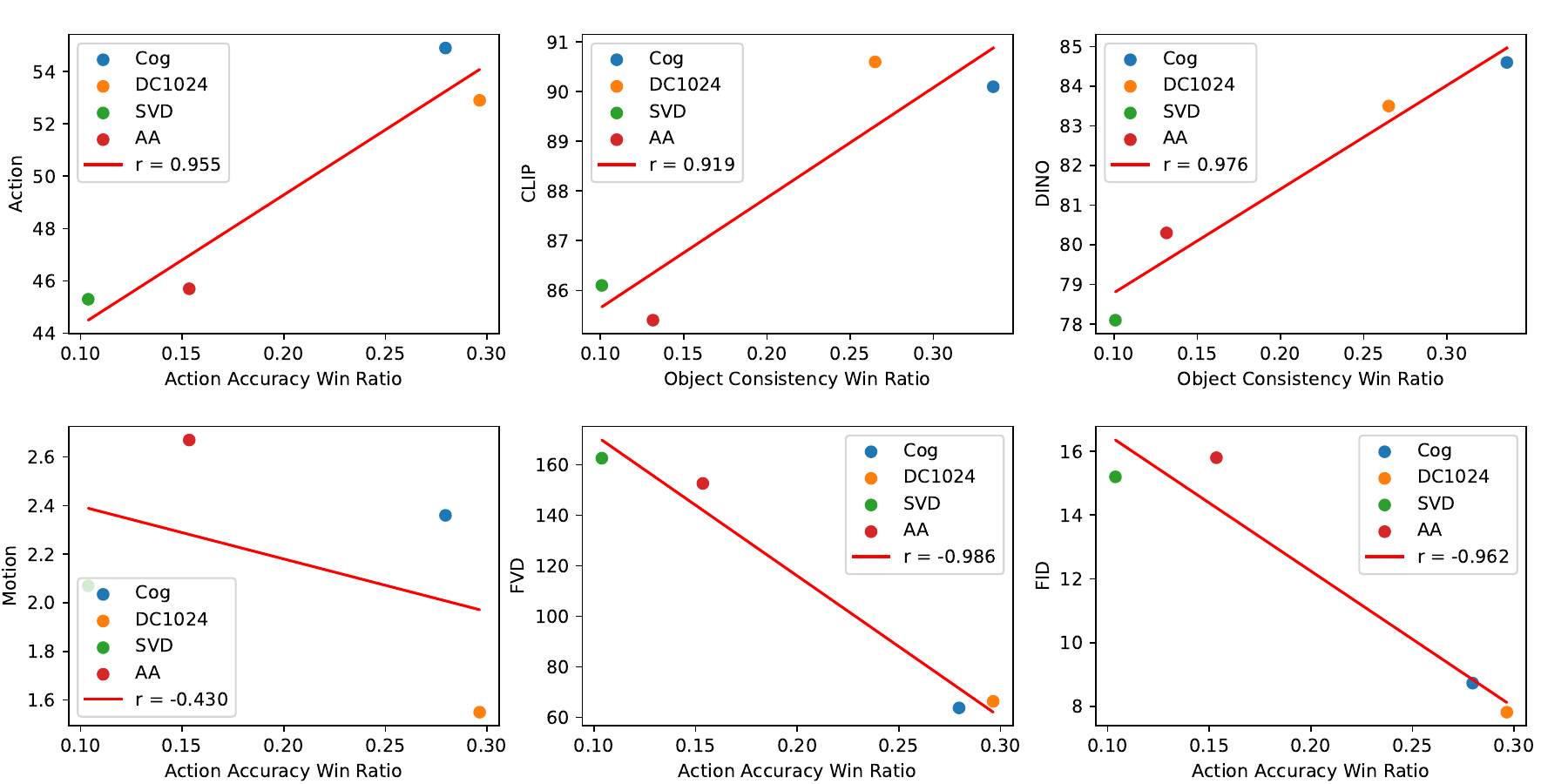}
   \caption{Correlation between automatic metrics and human evaluations for action accuracy and object consistency.}
   \label{fig:user}
\end{figure}

We then analyzed the correlation between human evaluations and several automatic metrics, including Action, FID, FVD, and Motion for action accuracy, as well as CLIP and DINO for object consistency. The results, illustrated in Figure~\ref{fig:user}, demonstrate a strong correlation between these automatic metrics and human evaluations, indicating that the metrics effectively capture human preferences in video generation tasks. This alignment suggests that the evaluated metrics can serve as reliable proxies for human evaluations in assessing video quality.

\section{Using MLLMs as a step planner}
\label{appendix llm}
As described in the main paper, we use different MLLMs to generate a sequence of steps, including detailed descriptions of each step. We present the versions of the MLLMs we used in Table~\ref{tab:llm version}.

\begin{table}[h]
\centering
\resizebox{1.\linewidth}{!}{
\begin{tabular}{lccc}
\hline
\textbf{Method}            & \textbf{SR$\uparrow$} & \textbf{Acc$\uparrow$} & \textbf{Version}\\ \hline
GPT-4o          & 11.3        & 28.5      &gpt-4o-2024-05-13   \\
GPT-4o-mini     & 1.80         & 15.4    &gpt-4o-mini-2024-07-18    \\
Claude 3.5 Sonnet& 15.6       & 33.2     &claude-3-5-sonnet-2024062   \\
ChatGPT-4o-latest  & 17.1       & 37.7 &gpt-4o-latest       \\
ChatGPT-4o-latest +RAG                    &\textbf{50.5}       & \textbf{65.0}  &-      \\ \hline
\end{tabular}}
\caption{Ablations on MLLMs planning. We present the versions of the MLLMs used in our study.}
\label{tab:llm version}
\end{table}

The detailed prompt is as follows. The [step pool] and [T] are only applicable in a close-set setting and will be removed in an open-set setting. The [examples] are the examples we obtained through retrieval argument.

\begin{quote}
\textbf{System:}

You are a helpful assistant, an expert in answering questions about action planning in instructional videos. Based on the given goal and picture, you need to infer the steps to accomplish this goal based on the current picture.

You have a pool of steps to choose from, and you should select the appropriate ones to describe the process. The step pool is [step\_pool].

Output the name of the steps and describe each step based on the given image. The description must be limited to one sentence.

Example: 1. pump up the fuel: A person is cutting a piece of wood, pumping up the fuel.

Example: 2. open the car key cover: A person holds a small black object, likely the car key, in their hand as they open the car key cover.

Example: 3. unscrew the screw and take out the needle: A person is carefully unscrewing the screw and taking out the needle from the sewing machine.

\textbf{User:}

Describe the steps to accomplish the goal based on the input image. The steps must be steps from the given step pool.

The goal is [goal];The number of steps is [T]; The initial image is [Image encoded with base64]; The example step sequences: [examples].

\textbf{Assistant:}

1. open the car key cover: A person is holding a car key and using a small tool to pry open the car key cover.

2. take out the car key battery: The car key cover is open, and the person is removing the old battery from its compartment.

3. put in the battery: The person places a new battery into the car key, ensuring it fits correctly in the compartment.

4. close the car key cover: The person closes the car key cover, securing it back into place.
\end{quote}

\section{Key-step image generation}
\label{appendix kig}

\textbf{Baselines.} We selected three open-source image generation models as baselines for comparison. The hyperparameters used for evaluating each model are provided in the Table~\ref{tab:imagesett}. For the Kolors~\cite{kolors} and SDXL~\cite{sdxl} models, we also explored fine-tuning, and the configurations used for fine-tuning are detailed accordingly.

\begin{table}[h]
\centering
\resizebox{1\linewidth}{!}{
\begin{tabular}{lccc}
\hline
\textbf{Hyperparameter} & \textbf{GenHowTo}~\cite{genhowto} & \textbf{Kolors}~\cite{kolors} & \textbf{SDXL}~\cite{sdxl} \\ 
\hline
Weight & GenHowTo-ACTIONS & Kolors-IP-Adapter-Plus & SDXL-base-IP-Adapter-Plus \\
Resolution & $512 \times 512$ & $768 \times 1344$ & $768 \times 1344$ \\
Sampler & DDIM & DPM & EDM \\
Steps & 50 & 50 & 50 \\
Guidance scale & 9 & 4 & 4 \\
IP-Adapter scale & - & 1 & 1 \\
Learnable blocks & - & \multicolumn{2}{c}{All Attention blocks in UNet (1.3B)} \\
Learning rate & - & $1 \times 10^{-6}$ & $1 \times 10^{-6}$ \\ 
Training steps & - & 5K & 5K \\
\hline
\end{tabular}}
\caption{Hyperparameters for baseline image models.}
\label{tab:imagesett}
\end{table}

\textbf{Skill Transformer.} During the training of the Skill Transformer, we employ Teacher Forcing to encourage faster convergence and stable learning. To enhance the consistency between the predicted features and the initial image features, we utilize a two-part loss function, both based on Mean Squared Error (MSE). The first loss measures the discrepancy between the predicted image features $f'_n$ and the ground-truth image features $f_n$. The second loss ensures consistency by assessing the difference between the predicted features $f'_n$ and the initial image features $f_0$. To balance the influence of these two losses, we multiply the second loss by a consistency weight.

During inference, we similarly combine the predicted features $f'_n$ and the initial image features $f_0$ using a fusion weight $w$. Specifically, we compute the final feature $\hat{f_n}$ as:  

\[
\hat{f_n} = w \cdot f'_n + (1-w) \cdot f_0
\]

The hyperparameters used, including the consistency weight and fusion weight, are detailed in Table~\ref{tab:stsett}. The fused image features are then injected into the image generation model through the IP-Adapter~\cite{ip} to synthesize the corresponding step images. The image generation process adopts the same hyperparameter configuration as the Baseline models.

\begin{table}[h]
\centering
\resizebox{.9\linewidth}{!}{
\begin{tabular}{lcc}
\hline
\textbf{Hyperparameter} & \textbf{ST for SDXL~\cite{sdxl}} & \textbf{ST for Kolors~\cite{kolors}} \\
\hline
Text encoder & CLIP-ViT-H-14 & CLIP-ViT-L-14-336 \\
Transfomer depth & \multicolumn{2}{c}{4} \\
Channels & \multicolumn{2}{c}{2048} \\ 
Head channels & \multicolumn{2}{c}{128} \\ 
Training steps & \multicolumn{2}{c}{10K} \\ 
Learning rate & \multicolumn{2}{c}{$1 \times 10^{-4}$} \\ 
Overall batch size & \multicolumn{2}{c}{32} \\ 
Learnable param. & \multicolumn{2}{c}{270M} \\ 
Consistency weight & \multicolumn{2}{c}{0.5} \\ 
Fusion weight & \multicolumn{2}{c}{0.5} \\
\hline
\end{tabular}}
\caption{Hyperparameters for Skill Transformer.}
\label{tab:stsett}
\end{table}

\section{Key-step clips generation}
\label{appendix kcg}

We fine-tune the AnimateAnything (AA)~\cite{aa}, Stable Video Diffusion (SVD)~\cite{SVD}, and DynamiCrafter (DC)~\cite{dynamicrafter} models. For the DC model, we test two variants: DC$_{512}$ and DC$_{1024}$. Additionally, we evaluate the zero-shot performance of the CogVideoX-5b-I2V~\cite{cogvideox} model. The hyperparameters used for evaluating each model are provided in the Table~\ref{tab:kcgsett}.

\begin{table}[h]
\centering
\resizebox{1\linewidth}{!}{
\begin{tabular}{lccccc}
\hline
\textbf{Hyperparameter} & \textbf{AA}~\cite{aa} & \textbf{SVD}~\cite{SVD} & \textbf{DC$_{512}$}~\cite{dynamicrafter} & \textbf{DC$_{1024}$}~\cite{dynamicrafter} & \textbf{CogVideoX}~\cite{cogvideox} \\
\hline
Resolution & $256 \times 384$ & $256 \times 384$ & $320 \times 512$ & $576 \times 1024$ & $480 \times 720$ \\
Inference Time & 0.14 & 0.12 & 0.41 & 1.55 & 3.29 \\
Sampler & DPM & EDM & \multicolumn{2}{c}{DDIM} & DPM \\
Steps & 25 & 25 & \multicolumn{2}{c}{50} & 25 \\
Guidance scale & 9 & 1.0-3.0 & \multicolumn{2}{c}{2} & 1.5 \\
Motion/FPS & 4 & 127 & \multicolumn{2}{c}{15} & - \\
Learnable param. & 641M & 669M & \multicolumn{2}{c}{315M} & - \\
Training steps & \multicolumn{4}{c}{5K} & - \\
Learning rate & \multicolumn{4}{c}{$5 \times 10^{-5}$} & - \\
Overall batch size & \multicolumn{4}{c}{16} & - \\
\hline
\end{tabular}}
\caption{Hyperparameters for Video Generator.}
\label{tab:kcgsett}
\end{table}

\section{More visualizations}
\label{appendix vis}

\begin{figure}[h]
  \centering
   \includegraphics[width=1.0\linewidth]{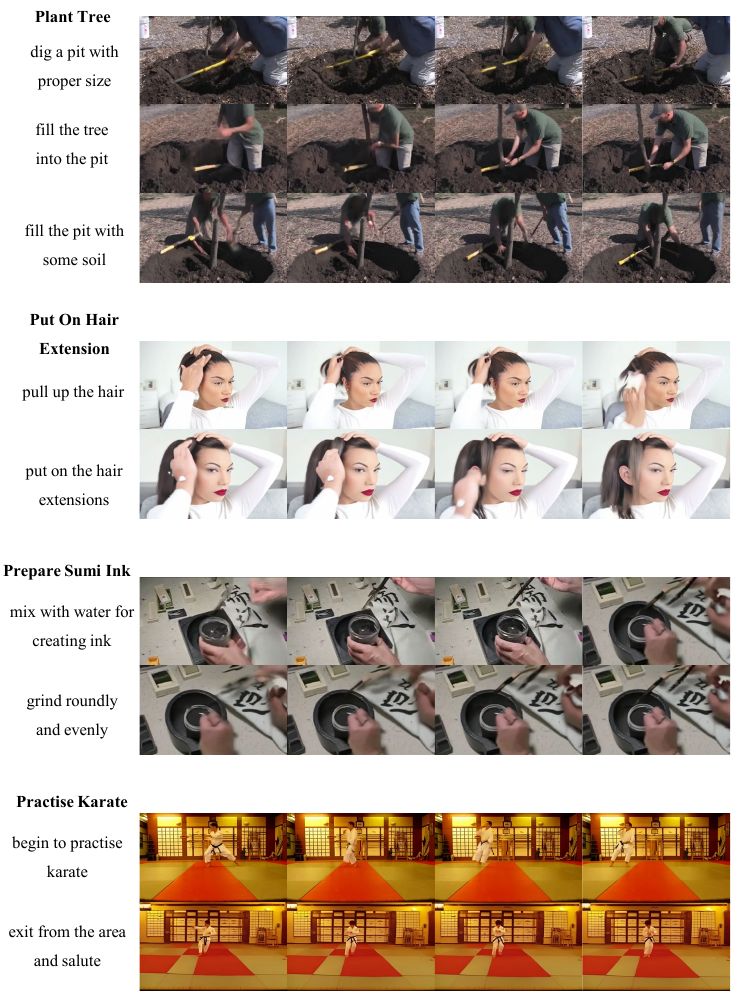}
   \caption{The visualization of the skill generator.}
   \label{fig:vis sup}
\end{figure}

\begin{figure}[t]  
  \centering
   \includegraphics[width=1.0\linewidth]{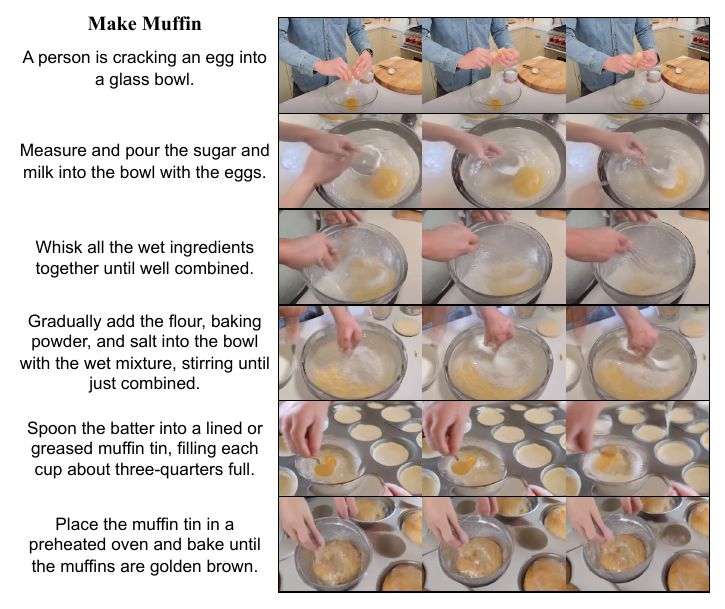}
   \caption{An example of an open-set complex skill.}
   \label{fig:rebuttal}
\end{figure}

We present additional visualizations with three skills in Figure~\ref{fig:vis sup}. In Figure~\ref{fig:rebuttal}, we present a complex skill example under the open-set setting and provide our complete step descriptions generated by the MLLM.

{
    \small
    \bibliographystyle{ieeenat_fullname}
    \bibliography{main}
}

% WARNING: do not forget to delete the supplementary pages from your submission 
% \input{sec/X_suppl}

\end{document}